\documentclass[11pt]{article}

\usepackage{geometry}
\usepackage{pdflscape}
\usepackage{booktabs}
\usepackage{tabularx}
\usepackage{threeparttable}
\usepackage{array}
\usepackage{setspace}
\usepackage{caption}

\newcolumntype{P}[1]{>{\raggedright\arraybackslash}p{#1}}
\newcolumntype{C}[1]{>{\centering\arraybackslash}p{#1}}
\newcolumntype{Y}{>{\raggedright\arraybackslash}X}

\usepackage[utf8]{inputenc}
\usepackage[T1]{fontenc}
\usepackage[english]{babel}

\usepackage{amsmath,amssymb,bm}
\usepackage{graphicx}
\usepackage[table]{xcolor}
\usepackage{booktabs}
\usepackage{tabularx}
\usepackage{array}
\usepackage{longtable}
\usepackage{setspace}
\usepackage{titlesec}
\usepackage{enumitem}
\usepackage{xurl}
\usepackage{csquotes}
\usepackage{threeparttable}
\usepackage[font=small,labelfont=bf]{caption}
\usepackage[hidelinks]{hyperref}
\usepackage[backend=biber,style=apa,natbib=true,doi=true,url=false,isbn=false,eprint=false,maxcitenames=2,maxbibnames=30]{biblatex}
\DeclareLanguageMapping{english}{english-apa}
\titleformat{\section}{\normalfont\large\bfseries}{\thesection}{0.6em}{}
\titleformat{\subsection}{\normalfont\normalsize\bfseries}{\thesubsection}{0.6em}{}
\titleformat{\subsubsection}{\normalfont\normalsize\itshape}{\thesubsubsection}{0.6em}{}
\newcolumntype{Y}{>{\raggedright\arraybackslash}X}
\newcolumntype{P}[1]{>{\raggedright\arraybackslash}p{#1}}
\newcolumntype{C}[1]{>{\centering\arraybackslash}p{#1}}

\newcommand{\1}{\mathbf{1}}
\definecolor{HeaderBlue}{RGB}{222,233,246}
\definecolor{HeaderGreen}{RGB}{226,239,218}
\definecolor{RowGray}{RGB}{248,249,250}
\title{\vspace{-3em}\textbf{Scalable and Trustworthy Earth Observation Foundation Models}}

\author{
Syed Usama Imtiaz\textsuperscript{1},
Mitra Nasr Azadani\textsuperscript{1},
and Nasrin Alamdari\textsuperscript{1}\\
\textsuperscript{1}Department of Civil and Environmental Engineering, Florida State University\\
Corresponding author: \textit{nalamdari@eng.famu.fsu.edu}
}

\date{}

\begin{document}
\maketitle
\vspace*{-1.5em}

\noindent\textbf{Abstract.}
Foundation models (FMs) have transformed machine learning from isolated task-specific model development toward general-purpose models pretrained on broad data and adapted to multiple downstream tasks. Earth observation (EO) is an important domain for this paradigm because satellite and airborne archives are large, high-revisit, and increasingly multimodal, while reliable field labels are often sparse. Remote sensing foundation models (RSFMs) cannot be transferred reliably/optimally without domain-specific adaptation. This is because EO data are governed by measurement physics and operational decision constraints. This chapter reviews the design principles arising from these domain-specific constraints. It first defines the FMs paradigm in remote sensing (RS), then synthesizes the current model landscape, pretraining objectives, architecture designs, downstream adaptation and trustworthiness requirements. The chapter also incorporates recent benchmark evidence showing that no single geospatial foundation model is universally best and that inconsistent evaluation remains a major issue to fair comparison and reliable deployment. In addition, two brief environmental monitoring case studies; physics-informed spectral targeted masking for harmful algal bloom prediction and reinforcement learning for adaptive environmental monitoring station selection to illustrate the FMs domain-guided principles in practice. This chapter posits that next-generation RSFMs should be evaluated not only by benchmark accuracy, but also by modality-aware transfer and physically plausible representations for trustworthy EO decisions.

\vspace{0.5em}
\noindent\textbf{Keywords:} remote sensing foundation models; earth observation; self-supervised learning; hyperspectral imagery; multimodal fusion; trustworthy AI.

\section{Introduction}

A \emph{foundation model} (FMs) is a model trained on broad data, usually by large-scale self-supervised learning (SSL) or weakly supervised learning, that can be adapted to many downstream tasks \citep{bommasani2021}. This concept is important because it changes the traditional workflow of machine learning model development. In supervised learning, a model is trained against defined labels, which therefore requires task-specific labeled data and task-specific training or adaptation, whereas SSL first learns a general transferable representation from large unlabeled data archives and then adapt that representation through approaches such as linear probing, prompting, full fine-tuning, or parameter-efficient tuning. This is highly advantageous for remote sensing (RS) because Earth observation (EO) archives are vast, high spatiotemporal frequency, and globally distributed, while expert labels and in situ measurements are expensive (delayed, unevenly sampled, and often tied to specific field campaigns). RSFMs now include visual FMs for imagery and time series, vision-language models for retrieval and visual question answering, generative models for synthesis and gap filling, multimodal systems for optical-radar-elevation fusion, large-language-model interfaces for geospatial reasoning, and emerging responsible-AI studies for EO \citep{xiao2025survey}. This breadth of modalities makes the field highly promising; however, it also creates a risk of RSFMs being defined merely as larger versions of ordinary vision models. This definition is conceptually incomplete because satellite data constitute a distinct machine learning modality and data: they are georeferenced, sensor-dependent, physically measured, and feed operationally consequential decisions \citep{rolf2024missioncritical}.

This chapter, therefore, posits RSFMs's design and evaluation around the properties of EO itself. These relevant properties include sensor modality, spectral band structure, spatial resolution, revisit schedule, atmospheric correction, viewing geometry, geolocation, physical process context, and deployment uncertainty. For instance, a model pretrained on RGB web images may learn useful texture or object shape, but it does not capture water-leaving reflectance, topographic shadow, or sensor reprocessing. The core objective of this chapter is to match model architecture, pretraining data, adaptation protocols, and evaluation design to the target EO problem. To illustrate this further, two environmental monitoring examples are included because they expose the practical consequences of RSFMs design: scarce labels, indirect targets, product-version dependence, uncertainty, and decisions under budget constraints. These examples strengthen RSFMs scientifically to learn representations for downstream tasks that rely on sensor physics and operational constraints.

\section{Key Definitions}
Several terms are useful before discussing RSFMs. A \emph{modality} is a type of measurement produced by a sensor and physical sensing principle, such as optical RGB imagery, multispectral imagery, hyperspectral imagery, synthetic aperture radar (SAR), thermal infrared imagery, lidar point clouds, or digital surface models. \emph{Spatial resolution} is the ground area represented by a pixel or measurement. \emph{Spectral resolution} describes how finely a sensor samples wavelength. \emph{Temporal resolution} describes revisit frequency and timing of observations. \emph{Geolocation} ties each observation to Earth coordinates. These properties are not metadata afterthoughts; they determine what information exists in the data and which model assumptions are valid.

\section{Remote sensing data require specialized foundation models}

RS is distinct from conventional computer vision; it is not equivalent to a larger-scale application of natural image models. While a natural image typically records reflected visible light from a human-centric perspective, a RS observation captures diverse physical phenomena (e.g., surface reflectance, microwave backscatter, emitted thermal radiation, canopy height, water color, or elevation) acquired across diverse platforms (i.e. satellite, airborne, and ground-based). These data vary significantly in their underlying physical units, noise profiles, spatial support, and temporal sampling schedules. In addition, modality diversity, spatial and spectral resolution, temporal dynamics, and domain discrepancy from natural images are central challenges for EO FMs \citep{xiao2025survey}.

EO differs from natural-image vision in at least five aspects. First, it is \emph{multimodal}. Optical data measure reflected solar radiation, SAR actively illuminates the surface with microwaves, lidar records three-dimensional structure, and thermal sensors measure emitted radiation. Second, EO is \emph{multi-resolution}. A high-resolution aerial pixel may represent part of a roof, while an ocean-color pixel may aggregate a broad water area. Third, EO is \emph{spectrally rich}. Hyperspectral sensors measure narrow contiguous bands, and neighboring wavelengths may contain physically interpretable absorption or scattering features. Fourth, EO is \emph{temporal}. Crops grow, floods rise and recede, forests burn and recover, and algal blooms intensify and transport. Fifth, EO is \emph{georeferenced}. Location improves contextual inference, but it can also encode geographic bias or enable privacy-sensitive inferences. These differences necessitate specialized RSFMs design. Random image crops can remove geospatial context. RGB-only pretraining can discard non-visible bands, and temporal shuffling can erase phenology or disturbance signals. A foundation model trained on one sensor may not transfer to another if wavelength, resolution, or atmospheric correction changes. Consequently, EO-oriented RSFMs increasingly include wavelength encodings, ground-sample-distance encodings, time and location embeddings, sensor-specific patch embeddings, multimodal fusion modules, and physically meaningful evaluation splits.

\begin{figure}[h]
    \centering
    \includegraphics[width=\textwidth]{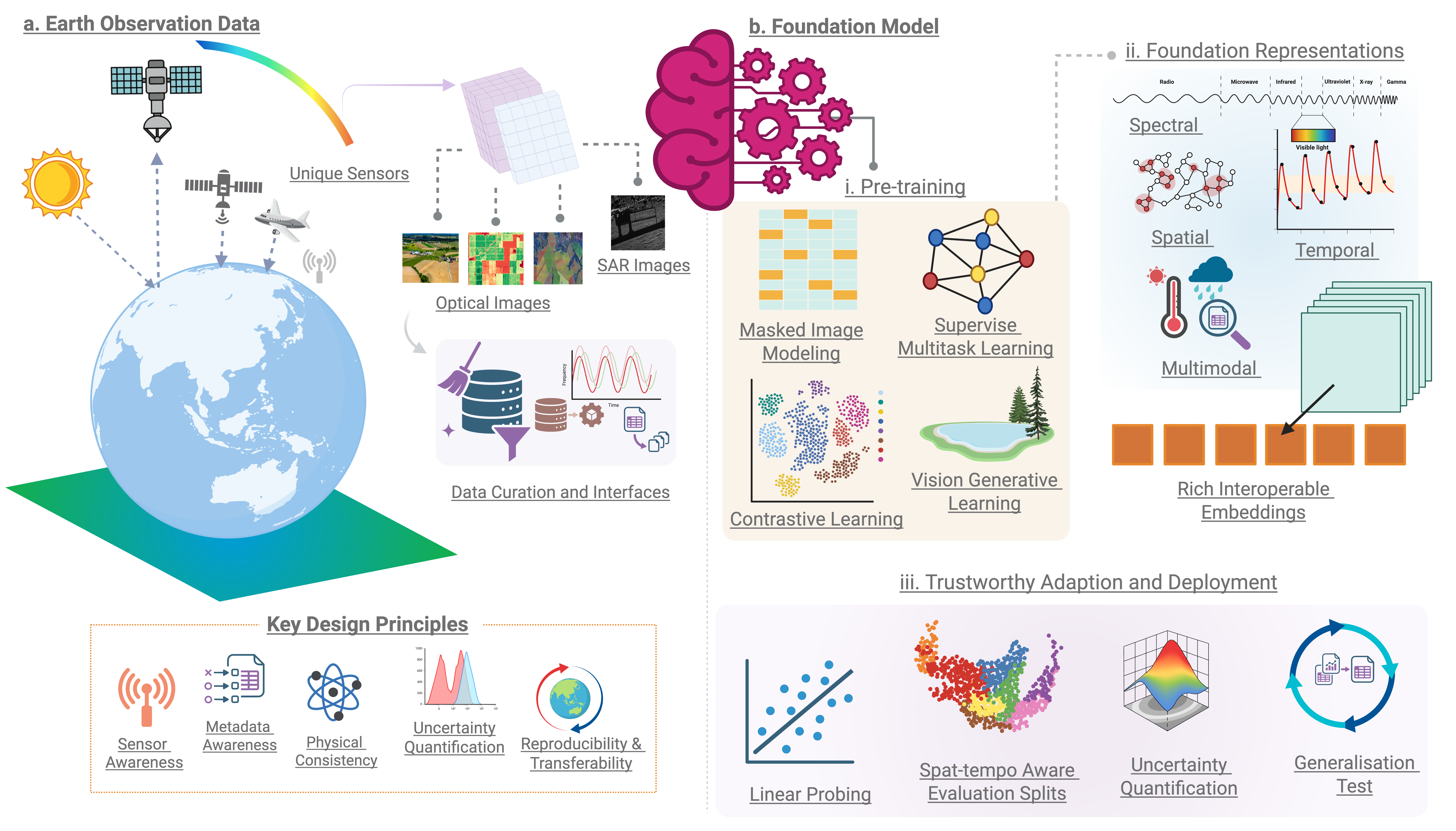}
    \caption{\textbf{Conceptual workflow for developing trustworthy Earth-observation foundation models (FMs).}
    \textbf{a.} Earth-observation (EO) data are acquired from heterogeneous platforms and sensors, including satellite, airborne and auxiliary observing systems, yielding multi-sensor optical and synthetic-aperture radar (SAR) image archives. Data-curation and interface layers harmonize imagery, metadata and derived products into analysis-ready inputs. Key design principles include sensor awareness, metadata awareness, physical consistency, uncertainty quantification, and reproducibility and transferability across sensors, regions and downstream tasks.
    \textbf{b.} FMs development incorporates large-scale pre-training objectives such as masked image modelling, supervised multitask learning, contrastive learning and vision-generative learning. These objectives produce rich, interoperable representations that capture spectral, spatial, temporal and multimodal structure in EO data. Trustworthy adaptation and deployment are supported through linear probing, spatio-temporally aware evaluation splits, uncertainty quantification and generalization testing.}
    \label{fig:eo_foundation_model_workflow}
\end{figure}

\section{EO Task Landscape} 

RSFMs are evaluated across an expansive spectrum of downstream tasks, and each task challenges distinct dimensions of the learned latent representation. For instance, scene classification requires a model to assign a singular categorical label to a discrete image patch that include classes such as residential, forest, airport, or cropland. Semantic segmentation assigns a thematic class to every constituent pixel and therefore rigorously evaluates spatial accuracy, boundary definition, and multiscale contextual understanding. Object detection localizes features including buildings, vehicles, aircraft, ships, or other discrete targets often through substantial variance in target scale and spatial orientation. Moreover, change detection systematically compares multi-temporal observations and evaluates whether an architecture can successfully differentiate true surface variance from confounding illumination or sensor-induced discrepancies. Visual question answering, captioning, and phrase grounding integrate natural language processing that demand precise semantic alignment between visual evidence and human textual queries \citep{xiao2025survey,li2024vrsbench}. 

Environmental monitoring introduces an entirely distinct category of tasks that is significantly less prominent in conventional computer vision benchmarks. It is the continuous biophysical parameter retrieval and operational decision support. Some relevant examples include quantifying chlorophyll levels, biomass volume, inundation depth, soil moisture, crop yield, toxin concentration, water temperature, or nutrient-related ecological risk. These tasks are inherently structured as continuous regression or temporal predictive problems rather than discrete categorization paradigms. In addition to these inherent complexities, they are also typically constrained by sparse, in situ measurements and regulatory or scientific thresholds. The first clearest example is harmful algal bloom (HAB) detection model, where model, may be evaluated not merely by root-mean-square error (RMSE) but by its statistical sensitivity in identifying critical anomalies exceeding an administrative management threshold to prevent human exposure. Second, an operational flood mapping model may be validated based on its capacity to reliably support actionable road closure or emergency evacuation decisions. Similarly, an agricultural crop model demands robust cross-year and cross-region out-of-distribution validity rather than high performance on independently and identically distributed (i.i.d.) randomly sampled pixels, which may overestimate accuracy because of spatial autocorrelation 

This structural diversity among downstream tasks directly motivates frameworks advocated by current benchmarking initiatives, shifting evaluation metrics to focus on isolating specific capability profiles rather than identifying a globally superior architecture. This approach requires testing the target according to precise domain needs (e.g., high-resolution spatial reasoning, multispectral and temporal dynamics, SAR robustness, cross-modal fusion, forecasting, language grounding, or label-scarce regression)\citep{simumba2025geobench2}. An architecture pretrained predominantly on high-spatial-resolution RGB imagery may exhibit strong performance in object detection but remain suboptimal for hyperspectral water-quality retrieval. Conversely, a model optimized on multispectral time series may excel at crop classification but prove less appropriate for generative aerial image captioning. Therefore, downstream tasks serve as a critical conceptual bridge that links model architecture design with empirical evaluation. Hence, RSFMs workflow (i.e. self-supervised pretraining objectives, architectural choices, and downstream adaptation strategies) requires careful selections that align with the downstream task.

To elaborate this further, task-specific evaluation explicitly deconstructs and offers a precise form of generalization claimed by a given framework. For instance, crop classification cross-validation evaluation varies based on year and region. This is because environmental phenology and agricultural management practices vary across space and time. Vision-language architectures require spatial grounding accuracy and systematic resistance to hallucinating erroneous geographic details.  These analytical distinctions prevent RSFMs from being treated as a homogeneous model. A selective, RSFMs evaluation does not require the exhaustive enumeration of every benchmark dataset result, but it must rigorously explicate whether a given model demonstrates utility for downstream task (e.g. perception, retrieval, forecasting, physical regression, or decision support). Table~\ref{tab:landscape} summarizes representative RSFMs and benchmark efforts, with emphasis on design choices for improved scientific transfer and evaluation interpretation. 


\begin{itemize}
    \item Masked and spectral pretraining strategies include SatMAE, Scale-MAE, CROMA, and SpectralGPT \citep{cong2022satmae,reed2023scalemae,fuller2023croma,hong2024spectralgpt}
    \item For multimodal or any-sensor strategies such as DOFA, AnySat, Copernicus-FMs, Prithvi-EO-2.0, Galileo, TerraMind, SkySense V2, and AlphaEarth \citep{xiong2024dofa,astruc2024anysat,wang2025copernicusfm,szwarcman2024prithvieo2,tseng2025galileo,jakubik2025terramind,zhang2025skysensev2,brown2025alphaearth}
    \item For vision-language, benchmark, and audit efforts such as RemoteCLIP, RS5M/GeoRSCLIP, VRSBench, EarthGPT, GeoLangBind, PANGAEA, GEO-Bench-2, MMEarth-Bench \citep{liu2024remoteclip,zhang2023rs5m,li2024vrsbench,zhang2024earthgpt,xiong2025geolangbind,marsocci2024pangaea,simumba2025geobench2,gordon2026mmearthbench,corley2026stateoftheart}
    \item Data infrastructure include SSL4EO-S12 and SSL4EO-L support self-supervised pretraining for Sentinel and Landsat imagery, respectively, while MMEarth and Major TOM support multimodal and interoperable EO data resources \citep{wang2023ssl4eo,stewart2023ssl4eol,nedungadi2024mmearth,francis2024majortom}.
\end{itemize}

\clearpage
\newgeometry{margin=1.8cm}
\begin{landscape}
\thispagestyle{empty}
\centering
\footnotesize
\setlength{\tabcolsep}{3.0pt}
\renewcommand{\arraystretch}{1.0}

\begin{threeparttable}
\captionof{table}{\textbf{Representative RSFMs model families, design requirements, and key lessons.}}
\label{tab:landscape}

\begin{tabularx}{\linewidth}{@{}
P{0.205\linewidth}
C{0.055\linewidth}
C{0.055\linewidth}
C{0.060\linewidth}
C{0.060\linewidth}
C{0.055\linewidth}
Y
@{}}
\toprule
\textbf{Model family} &
\textbf{Sensor\newline bands} &
\textbf{Multi.} &
\textbf{Time/\newline place} &
\textbf{Text/\newline gen.} &
\textbf{Eval.} &
\textbf{Selected examples and design lesson} \\
\midrule

Masked and spectral self-supervision\newline
\textit{SatMAE, Scale-MAE, SpectralGPT} &
x &  & x &  &  &
Masked pretraining is most useful when reconstruction reflects EO structure: temporal stacks, multiscale imagery, diagnostic wavelengths, or spatial--spectral dependencies rather than generic RGB patches \citep{cong2022satmae,reed2023scalemae,hong2024spectralgpt}. \\

\addlinespace[0.2em]

Cross-modal and multimodal SSL\newline
\textit{CROMA, Galileo, SkySense V2} &
x & x & x &  &  &
Optical, SAR, elevation, weather, and other modalities provide complementary evidence, but fusion must account for different resolutions, noise sources, revisit times, and physical sensing mechanisms \citep{fuller2023croma,tseng2025galileo,zhang2025skysensev2}. \\

\addlinespace[0.2em]

Sensor- and metadata-conditioned backbones\newline
\textit{DOFA, AnySat, Copernicus-FMs, Prithvi-EO-2.0} &
x & x & x &  &  &
Wavelength, sensor identity, scale, time, and location increasingly function as inputs to the model rather than as external bookkeeping variables \citep{xiong2024dofa,astruc2024anysat,wang2025copernicusfm,szwarcman2024prithvieo2}. \\

\addlinespace[0.2em]

Vision--language and grounded geospatial language models\newline
\textit{RemoteCLIP, RS5M/GeoRSCLIP, EarthGPT, GeoLangBind, GeoMeld, VRSBench} &
 & x & x & x & x &
Language interfaces support retrieval, captioning, question answering, and grounding; their scientific value depends on whether generated text is verified against image, sensor, and geographic evidence \citep{liu2024remoteclip,zhang2023rs5m,zhang2024earthgpt,xiong2025geolangbind,hasan2026geomeld,li2024vrsbench}. \\

\addlinespace[0.2em]

Generative, any-to-any, and temporal-flow EO models\newline
\textit{TerraMind, TerraFlow, HighFMs} &
x & x & x & x & x &
Generative and sequence-aware objectives can support gap filling, modality translation, temporal completion, high-frequency monitoring, and risk mapping, but generated or forecast outputs require physical and calibration checks \citep{jakubik2025terramind,puriy2026terraflow,girtsou2026highfm}. \\

\addlinespace[0.2em]

Benchmarks, sustainability, and domain-specific audits\newline
\textit{PANGAEA, GEO-Bench-2, MMEarth-Bench, SustainFMs, Cryo-Bench, GFMs audits} &
 & x & x &  & x &
Benchmark studies show that no single model should be treated as universally best; comparisons should be capability-specific and should report splits, task families, energy or compute cost, artifacts, variance, and deployment assumptions \citep{marsocci2024pangaea,simumba2025geobench2,gordon2026mmearthbench,ghamisi2025sustainfm,kaushik2026cryobench,corley2026stateoftheart}. \\

\bottomrule
\end{tabularx}

\vspace{0.4em}

\begin{minipage}{1.0\linewidth}
\footnotesize
\centering
\setstretch{1.0}
\textit{Abbreviations.} EO = Earth observation; GFMs = geospatial Foundation model; RSFMs = remote sensing Foundation model; SAR = synthetic aperture radar; SSL = self-supervised learning; Multi. = multimodal; gen. = generative output; Eval. = Evaluation.  ``x'' marks a primary design emphasis; blank cells indicate a secondary or indirect relationship. 
\textit{Reading guide.} The table compares model families by the RSFMs design requirements to broader design lessons about wavelength structure, multimodality, temporal context, language grounding, generation, sustainability, and benchmark design.
\end{minipage}

\end{threeparttable}

\vfill
\makebox[\linewidth][c]{\normalsize\thepage}
\end{landscape}
\restoregeometry
\clearpage

\subsection{Supervised and multitask pretraining}

Supervised pretraining uses labeled EO datasets to initialize a model backbone before downstream adaptation. Its advantage is semantic alignment (e.g. labels for land cover, objects, roads, buildings, crops, or segmentation masks) that can teach useful features. Multitask supervised pretraining expands this idea by learning from several label types at once. The limitation is that supervised pretraining depends on annotation availability and annotation quality, which can create model bias. Regions and tasks with abundant labels can dominate, while high-value scientific variables (such as toxin concentration, gene abundance, soil carbon) may remain poorly represented. For this reason, supervised pretraining should be treated as a useful baseline. It is strongest for common perception tasks, but it cannot by itself solve the label scarcity that motivates RSFMs for environmental science.

\subsection{Contrastive self-supervision}

Contrastive self-supervision learns by pulling related views together in an embedding space and pushing unrelated views apart. In EO, positive pairs can be physically meaningful. For instance, the same location across seasons, optical and SAR images of the same area, adjacent dates in a stable period, or co-located satellite and ground-level observations. A common objective is
\begin{equation}
\mathcal{L}_{\mathrm{NCE}}=-\frac{1}{B}\sum_{i=1}^{B}\log\frac{\exp(\mathbf{z}_i\cdot\mathbf{z}_i^+/\tau)}{\sum_{j=1}^{B}\exp(\mathbf{z}_i\cdot\mathbf{z}_j/\tau)},
\end{equation}
where $\mathbf{z}_i$ is an embedding, $\mathbf{z}_i^+$ is a positive view, and $\tau$ is the temperature. 

The strength of contrastive learning is invariance. However, excessive invariance can suppress scarce important labels. If the positive-pair definition treats a meaningful land-cover change, bloom event, or flood signal as a nuisance, the representation may suppress the process the downstream model needs to detect.

\subsection{Masked image and masked spectral modeling}

Masked image modeling trains a model to reconstruct missing patches or tokens from visible context \citep{he2022mae}. In RS, MIM has been adapted to temporal resolution, spatial resolution, orientation, small objects, and spectral dimension \citep{xiao2025survey,wang2022ssl}. For hyperspectral EO, masking is not just a technical choice. It defines what physical relationships the encoder is forced to learn. Uniform random masking treats all bands as equally informative, while environmental processes often have diagnostic wavelengths. A water-quality model may need to emphasize pigment-sensitive regions; a mineral model may need absorption features; a vegetation model may need red-edge and shortwave-infrared features. 

This problem motivates physics-informed masked spectral modeling. If $D$ is a set of diagnostic bands, a targeted mask can be written as
\begin{equation}
 m_b=\1[b\in D],
\end{equation}
with reconstruction loss applied only on those bands. The model is then trained to infer physically important wavelengths from spectral context \citep{gawrysiak2025phism}.

\subsection{Vision-language, vision-location, and generative objectives}

Vision-language models (VLMs) align EO imagery with text so that images can be retrieved, described, queried, or grounded through language. VLMs and related benchmark datasets, such as VRSBench, RS5M/GeoRSCLIP, EarthGPT, and GeoLangBind learn through captioning, grounding, visual question answering, instruction following, and language-mediated cross-sensor fusion \citep{xiao2025survey,li2024vrsbench,zhang2023rs5m,zhang2024earthgpt,xiong2025geolangbind}. In EO, VLMs align images with location metadata, making geospatial coordinates as part of representation learning. Generative objectives broaden this scope further. Diffusion and any-to-any multimodal models can synthesize missing observations, inpaint cloud gaps, translate between modalities, generate super-resolved imagery, or create auxiliary modalities during inference. TerraMind illustrates this direction by framing EO as multimodal generation rather than only feature extraction \citep{jakubik2025terramind}. However, generative EO should be evaluated for physical consistency, uncertainty, and  potential hallucination, especially when they are used for environmental monitoring or decision support.

\section{Architectures and Conditioning Mechanisms}

RSFMs analysis should distinguish between the backbone architecture and the EO interface.  The backbone is the main AI architecture (like a vision transformer or a convolutional neural network) that processes data. The EO interface is the gateway that allows satellite-specific information (such as sensor wavelengths, dates, resolution, and locations) to enter the model. Many weaknesses of generic vision transfer arise because the interface is wrong even if the backbone is powerful.

\subsection{Backbones and sensor interfaces}

Current RSFMs widely adopt transformer models. This is because token sequences can naturally represent image patches, spectral groups, time steps, language tokens, and metadata. Convolutional models remain useful for local spatial structure and efficient dense prediction, whereas hybrid models can combine local inductive bias with global attention. However, it is pertinent to note that a transformer alone does not make a model RS aware. The model also needs a mechanism to represent the wavelength, scale, modality, and time. In this regard, \citep{xiong2024dofa} provides a concrete example of wavelength-aware modeling in which a dynamic hypernetwork adjusts the model according to wavelengths so that one transformer can work across multiple sensors. \citep{szwarcman2024prithvieo2} provides a concrete example of temporal and location embeddings in large-scale HLS time-series pretraining. \citep{tseng2025galileo} illustrates flexible multimodal modeling with global and local self-supervised features . \citep{astruc2024anysat,wang2025copernicusfm} push the same design principle further by treating resolution, scale, sensor identity, and metadata as first-class input conditions rather than fixed preprocessing assumptions. \citep{brown2025alphaearth} reframes the output as a global embedding field that can support sparse-label mapping, showing another route from learned foundation representations to operational geospatial products.  

\subsection{Spectral and temporal tokenization}

Hyperspectral imagery requires preserving wavelength order and band identity. Let
$\mathbf{x}\in\mathbb{R}^B$ denote a per-pixel hyperspectral reflectance spectrum,
where $B$ is the number of spectral bands and $\mathbb{R}$ denotes the real-valued
space. The spectrum can be partitioned into $T$ spectral tokens:
\begin{equation}
\mathbf{u}_t=g_\theta(\mathbf{x}_{\mathcal{B}_t},\mathbf{e}_{\lambda,t}), 
\qquad t=1,\ldots,T,
\end{equation}
where $t$ indexes the spectral token, $\mathbf{u}_t$ is the learned token embedding,
$\mathcal{B}_t\subseteq\{1,\ldots,B\}$ is the set of band indices or wavelength interval
assigned to token $t$, and $\mathbf{x}_{\mathcal{B}_t}$ is the corresponding subvector
of reflectance values. The function $g_\theta(\cdot)$ is a learnable tokenization or
projection function with parameters $\theta$, $\mathbf{e}_{\lambda,t}$ is the wavelength
encoding for token $t$, and $\lambda$ denotes wavelength. 

This representation allows the model to learn cross-band covariance while retaining physical order. For environmental retrieval, input representations should be domain-informed to support the specific downstream task. EO time series are often irregular because of clouds, sensor gaps, changing revisit intervals, and compositing. A temporal encoder should know whether observations are daily, weekly, seasonal, or missing. For crops, cross-year evaluation is critical; for flood mapping, event-level holdout is more meaningful than random pixel splitting; for HAB monitoring, station-time and season-level holdouts better reflect deployment.

\subsection{Metadata conditioning and multimodal fusion}

RS observations include a wide array of metadata (such as acquisition date, sensor, wavelength, solar angle, viewing angle, cloud flag, atmospheric correction level, product version, ground sample distance, and geolocation). Conditioning layers can inject this information into a model for improved learning. A generic feature-wise modulation layer, following feature-wise linear modulation (FiLM) \citep{perez2018film}, is
\begin{equation}
\mathrm{FiLM}(\mathbf{F}\mid \mathbf{c})=\gamma(\mathbf{c})\odot\mathbf{F}+\beta(\mathbf{c}),
\end{equation}
where $\mathbf{F}$ is an intermediate representation, $\mathbf{c}$ is conditioning metadata, and $\gamma(\mathbf{c})$ and $\beta(\mathbf{c})$ are learned functions that generate feature-wise scale and shift parameters. The principle also supports wavelength-conditioned encoders, sensor-conditioned patch embeddings, and location-aware adaptation. 

Multimodal model have shown improved performance, however, feature fusion should be physically aligned. For instance, a SAR image, an optical reflectance map and weather data may have different resolution, support, and time. Fusion can occur early at the token level, intermediately through cross-attention, late at the prediction level, or selectively through mixture-of-experts routing. The most suitable choice depends on the scientific question. Agriculture and disaster response may benefit from optical-SAR fusion; water-quality retrieval may benefit from hyperspectral imagery combined with meteorology and hydrology; urban analysis may benefit from image-elevation fusion.

\section{Downstream Adaptation}

Adaptation is the stage where a foundation model performs a specific EO task. Common methods include linear probing, full fine-tuning, and parameter-efficient tuning. Linear probing freezes the encoder and trains a simple head to test the representation quality. Full fine-tuning updates all weights and can improve task performance but may overfit small labeled datasets. Moreover, parameter-efficient fine tuning (PEFT) methods such as adapters, LoRA, and bias-only tuning update a small number of parameters while preserving the base model \citep{houlsby2019adapters,hu2022lora,zaken2022bitfit}. These methods are especially relevant when a single RS must be adapted to many regions, seasons, and agencies. In this regard, recent geospatial PEFT studies demonstrate that adaptation costs, geographic generalization constraints, decoder configurations, and metadata integration collectively all influence the model's post-tuning operational utility and out-of-distribution robustness \citep{martiescofet2025peft}. FMs are not automatic replacements for smaller supervised, statistical, or physics-based models. FMs become most valuable when transfer, label efficiency, multimodal context, or repeated adaptation across many tasks is needed \citep{marsocci2024pangaea}. In the development of FMs, negative or mixed transfer results are scientifically valuable because they reveal when domain-specific data, physics-based models, or simpler supervised baselines remain preferable.

\section{Evaluation}
Evaluation of Geospatial Foundation model (GFMs)  has often been too narrow, and GFMs do not consistently outperform supervised baselines across diverse settings \citep{marsocci2024pangaea} because model choice depends on task, modality, resolution, and constraints. GEO-Bench-2 \citep{simumba2025geobench2} highlighted that evaluation shift from one aggregate score toward different FMs groups Table~\ref{tab:landscape} and reported that no single model dominates across all tasks. MMEarth-Bench extends this concern to globally distributed multimodal environmental tasks and shows that multimodal pretraining can help under limited labels while geographic generalization remains difficult \citep{gordon2026mmearthbench}. In brief, GFMs comparisons remain difficult because of inconsistent protocols, unreleased weights, unique pretraining configurations, and incomplete variance reporting \citep{corley2026stateoftheart}. The implication here is practical. For instance, a benchmark score for land-cover classification does not guarantee performance for HAB monitoring or a model that performs well on high-resolution RGB object detection may not be best for multispectral crop mapping. A model that reconstructs reflectance well may still be poorly calibrated for decision support. Model evaluation criteria should also align with the objective of the model.

\clearpage
\newgeometry{margin=1.8cm}
\begin{landscape}
\thispagestyle{empty}
\centering
\footnotesize
\setlength{\tabcolsep}{1.0pt}
\renewcommand{\arraystretch}{0.6}

\begin{threeparttable}
\captionof{table}{\textbf{Evaluation and reporting checks for RSFMs studies.}}
\label{tab:evaluation_checklist}

\begin{tabularx}{\linewidth}{@{}
P{0.100\linewidth}
C{0.070\linewidth}
C{0.070\linewidth}
C{0.075\linewidth}
C{0.075\linewidth}
C{0.065\linewidth}
Y
@{}}
\toprule
\textbf{Reporting element} &
\textbf{Transfer\newline split} &
\textbf{Product\newline trace} &
\textbf{Uncert.\newline calib.} &
\textbf{Dec.\newline utility} &
\textbf{Reuse\newline files} &
\textbf{Selected reporting requirement and failure mode addressed} \\
\midrule

Deployment split design &
x &  &  & x &  &
Report temporal, geographic, station, event, or sensor holdout matched to the intended use case; report split-specific confidence intervals or blocked bootstrap intervals to avoid inflated accuracy from spatial or temporal autocorrelation \citep{marsocci2024pangaea,simumba2025geobench2,gordon2026mmearthbench}. \\

\addlinespace[0.25em]

Sensor and product specification &
 & x &  &  & x &
Report sensor, processing level, product version, quality flags, atmospheric correction, compositing window, and satellite--field match-up tolerance. \\

\addlinespace[0.25em]

Adaptation protocol &
x &  &  &  & x &
Report linear probing, parameter-efficient tuning, full fine-tuning, label fraction, hyperparameters, seeds, model size, latency, and compute budget; this distinguishes representation quality from tuning effort and operational cost. \\

\addlinespace[0.25em]

Baseline and benchmark context &
x & x &  & x &  &
Compare with strong supervised, physics-based, task-specific, and simple statistical baselines using comparable inputs \citep{kaushik2026cryobench,girtsou2026highfm}. \\

\addlinespace[0.25em]

Uncertainty and calibration &
 &  & x & x &  &
Report reliability diagrams, expected calibration error (ECE), negative log-likelihood (NLL), interval coverage, conformal coverage when used, and continuous ranked probability score (CRPS) for probabilistic regression\citep{guo2017calibration,lakshminarayanan2017deepensembles}. \\

\addlinespace[0.25em]

Threshold and decision metrics &
x &  & x & x &  &
For advisory or event detection tasks, report threshold-specific recall, precision, false-alarm rate, Brier score, detection timeliness, regret, sampling cost, spatial coverage, and robustness to missing observations. Example Model TerraFlow \citep{puriy2026terraflow}. \\

\addlinespace[0.25em]

Language grounding and semantic verification &
 & x &  & x & x &
For VLM or MLLM systems, report grounding accuracy, object-level factuality, geographic hallucination checks, and caption-verification procedures. Example Model GeoMeld \citep{hasan2026geomeld}. \\

\addlinespace[0.25em]

Energy, carbon, and sustainability reporting &
 &  &  & x & x &
Report training and adaptation energy, approximate carbon footprint where feasible, GPU-hours, inference latency, and model-size trade-offs. Example Model SustainFMs \citep{ghamisi2025sustainfm}. \\

\addlinespace[0.25em]

Artifacts and reuse conditions &
 & x &  &  & x &
Report code, weights, model cards, data-access instructions, evaluation harnesses, licenses, known limitations, and exact commits \citep{corley2026stateoftheart}. \\

\bottomrule
\end{tabularx}

\vspace{0.1em}

\begin{minipage}{1.0\linewidth}
\footnotesize
\centering
\setstretch{1.0}
\textit{Abbreviations.} CRPS = continuous ranked probability score; ECE = expected calibration error; GFMs = geospatial Foundation model; MLLM = multimodal large language model; NLL = negative log-likelihood; RSFMs = remote sensing Foundation model; VLM = vision--language model. ``x'' marks a primary audit objective; blank cells indicate a secondary or indirect relationship.
\textit{Reading guide.} The columns separate concerns for evaluation in transfer design tests.
\end{minipage}

\end{threeparttable}

\vfill
\makebox[\linewidth][c]{\normalsize\thepage}
\end{landscape}
\restoregeometry
\clearpage

\section{Trustworthiness, Scientific Validity, and Reproducibility}

In environmental systems, trustworthy and faithful predictions are the essential of actionable science, therefore, trustworthiness in RSFMs is of paramount importance. 
This trust depends on the entire workflow. Here, trustworthiness refers to design choices that improve physical plausibility, and responsible development of these models must prioritize data privacy, security, fairness, and ethical use \citep{ghamisi2025responsible}, and model predictions must be physically plausible and interpretable. For instance, geographic representativeness is a major concern. EO datasets are often biased toward regions with more sensors, labels, cloud-free images. A model that performs well on global averages may still fail in under-monitored watersheds or areas with different atmospheric and optical conditions. Geographic bias is not only a statistical problem; it affects environmental justice when monitoring decisions influence advisories or disaster response. Moreover, RSFMs can fail under systematic or natural errors (e.g. clouds, haze, sun glint, atmospheric correction errors, sensor drift, shallow-water bottom reflectance, mixed pixels, seasonal shifts, or changed viewing geometry). Robustness tests may use physically realistic perturbations rather than only generic noise. Interpretability should be tied to sensor physics. In EO, explanations can be checked against known spectral and geophysical relationships. However, physical plausibility is not a causal proof. It should be described as evidence that a model uses meaningful measurement structure to learn known mechanistic behavior in making predictions. Today, reproducibility is a major weakness in the broader GFMs literature with inconsistent protocols and incomplete weight release supports a simple conclusion. RSFMs studies should provide enough information for readers to compare, repeat, and diagnose results \citep{corley2026stateoftheart}. Trustworthy FMs requires at minimum these five questions before deployment. 
\begin{enumerate}
    \item Is the training data representative of the intended deployment setting? 
    \item Are the learned features physically plausible for the sensor and target process? 
    \item Are uncertainty estimates calibrated enough to support action? 
    \item Are failure modes documented under sensor and environmental perturbations? 
    \item Can the obtained results be reproduced?
\end{enumerate}
\section{Application Examples in Environmental Monitoring}
The following examples are included to illustrate how RSFMs design principles can be applied in environmental monitoring. Each example is organized around the problem setting, model idea, reported evidence, limitations, and general lesson for RSFMs design.
\subsection{Case Study 1: SpecTM for physics-informed hyperspectral pretraining}
\begin{figure}[htbp]
\centering
\includegraphics[width=1.00\textwidth]{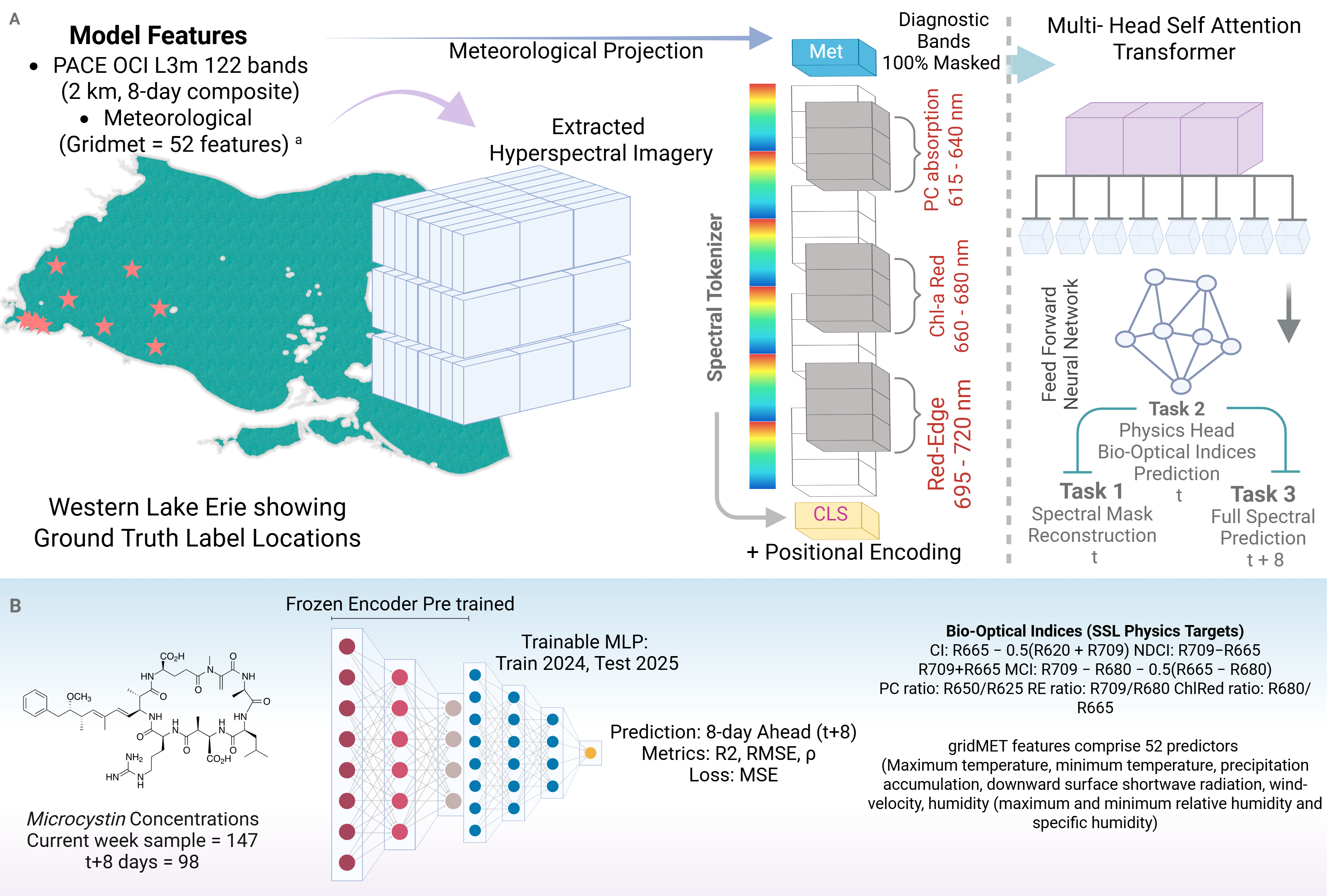}
\caption{The figure explains SpecTM workflow: Spectral tokens are constructed and augmented with positional encoding, while key diagnostic bands are masked to enforce physics-guided learning. The pretrained encoder is frozen and coupled with a trainable MLP head to predict microcystin concentrations 8 days ahead. The figure use is with permission from \citep{imtiaz2026spectm} }
\label{fig:spectm}
\end{figure}

\noindent\textbf{Problem.} HAB monitoring requires estimating conditions related to cyanobacteria and toxins from satellite data, but field measurements of microcystin are sparse. PACE's Ocean Color Instrument provides hyperspectral ocean-color observations that can resolve pigment-sensitive regions more finely than multispectral sensors \citep{werdell2019pace}. Phycocyanin, chlorophyll-related absorption, and red-edge behavior provide useful evidence for cyanobacteria, but microcystin itself is not directly observed from space \citep{mishra2013phycocyanin,mishra2021cyanobacteria,stumpf2016cyanotoxin}.

\noindent\textbf{Model idea and evidence.} SpecTM modifies masked spectral modeling by masking diagnostic spectral regions associated with pigment-sensitive bands rather than masking uniformly at random. The model jointly optimizes diagnostic-band reconstruction, bio-optical index prediction, and 8-day temporal prediction before fine-tuning on scarce microcystin labels \citep{imtiaz2026spectm}. Reported results include improved current-week and 8-day-ahead prediction, where SpecTM achieved \(R^2=0.695\) for current-week and \(R^2=0.620\) for 8-day-ahead microcystin prediction, outperforming the reported baseline models by 34\% and 99\%, respectively; targeted masking further improved performance by \(+0.037\) \(R^2\) over matched random masking.

\noindent\textbf{Limitations and general lesson.} It shows that a physically meaningful pretext task can improve representation learning when domain knowledge is encoded and downstream scarce-label prediction in the reported Lake Erie setting. External validation across different lakes, product versions, seasons, and optical water types remains necessary.

\subsection{Case Study 2: PiCSRL for adaptive station selection}

\begin{figure}[h]
\centering
\includegraphics[width=1.00\textwidth]{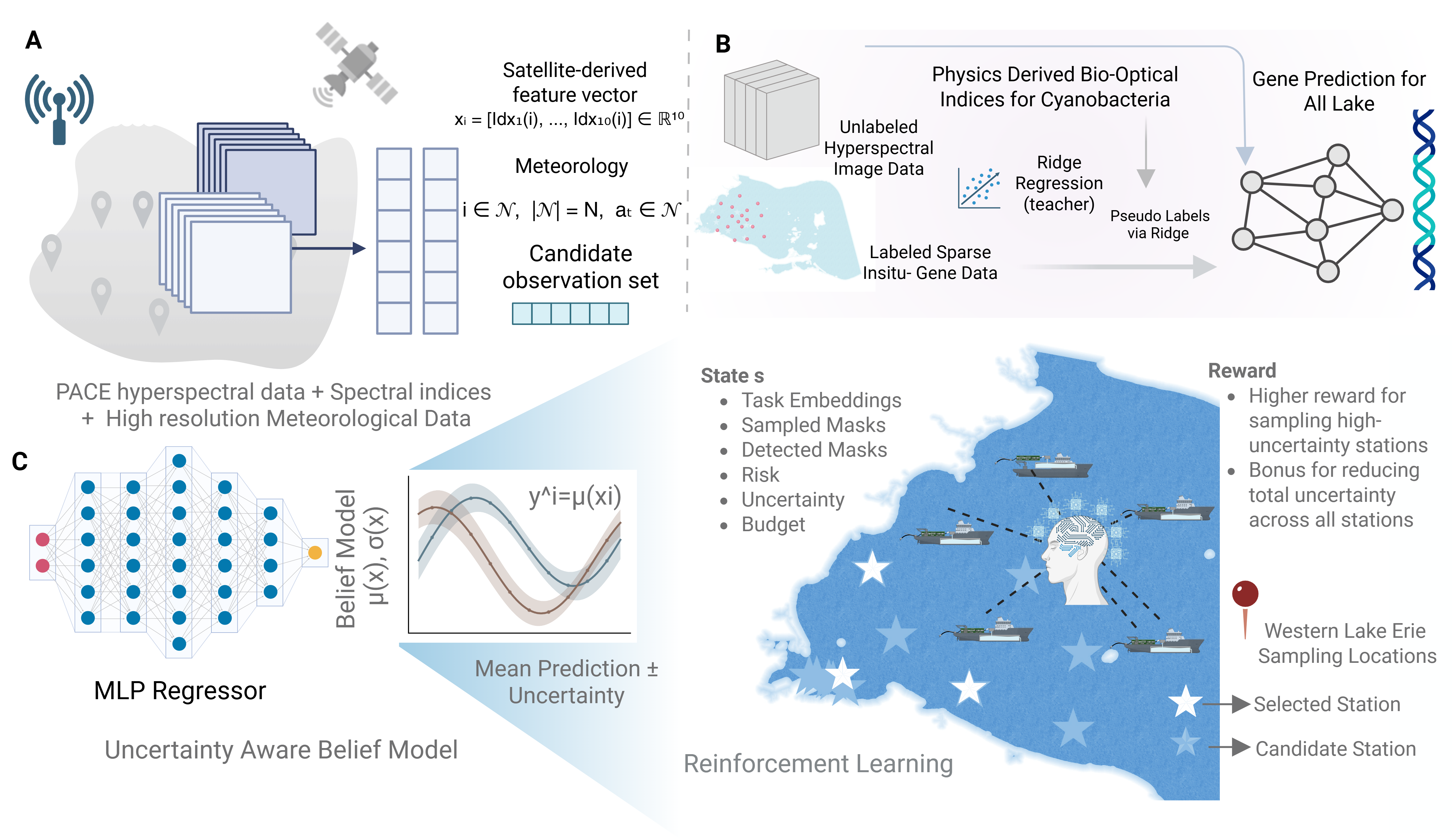}
\caption{The framework explains Physics-informed bio-optical indices and sparse in-situ observations are integrated through a semi-supervised learning framework and an uncertainty-aware belief model, and predictions are then used in a reduced RL state representation to guide adaptive station selection. The figure use is with permission from \citep{azadani2026picsrl}. }
\label{fig:picsrl}
\end{figure}

\noindent\textbf{Problem.} Environmental monitoring is often constrained by field-sampling budgets. In this setting, the model must support station-selection decisions rather than prediction alone. This is a sequential adaptive sensing problem rather than only a prediction problem.

\noindent\textbf{Model idea and evidence.} PiCSRL formulates cyanobacterial monitoring as budget-constrained station selection under high-dimensional, low-sample-size conditions \citep{azadani2026picsrl}. A compact representation function maps hyperspectral observations to physics-informed features; a belief model estimates predictive means and epistemic uncertainty; and a deep Q-learning policy selects stations from a state containing predicted values, uncertainty, visitation mask, and remaining budget \citep{mnih2015dqn,vanhasselt2016double}. Reported results show improved reconstruction error and bloom detection compared with random, greedy, and UCB-style baselines, with scalability analysis on a larger candidate network. For selecting \(K=3\) stations from \(N=8\) candidates, PiCSRL achieved lake-wide reconstruction \(RMSE=0.153\) and a 98.4\% bloom-detection rate, compared with \(RMSE=0.296\) for random selection and \(RMSE=0.178\) for the UCB baseline. The physics-informed representation also improved test generalization, with \(R^2=0.52\) using physics-informed indices compared with \(R^2=0.41\) using raw spectral bands.

\noindent\textbf{Evidence boundary and limitations.} PiCSRL results use engineered physics-informed bio-optical features and teacher-student semi-supervised belief modeling. Replacing or augmenting that representation with a pretrained SpecTM-style encoder is a promising future extension. This distinction is important for scientific correctness. Additional evaluation should include station holdout, budget sensitivity, stronger active-sensing baselines such as Gaussian-process upper-confidence-bound methods and information-directed sampling, and calibration diagnostics for the belief model \citep{srinivas2010gpucb,russo2018ids}. The general lesson is decision-aware representation learning. In conventional RSFMs evaluation, a representation is evaluated by predictive metrics. In adaptive sensing, it should also be evaluated by the quality of the decisions it enables: regret, detection timeliness, uncertainty reduction, spatial coverage, cost, and robustness when observations are missing or labels are delayed.

\section{Challenges and Future Outlook}

The next stage of RSFMs research should combine broad foundation-model capability with careful EO science. Larger datasets are necessary, but they are not sufficient without inclusion of metadata (e.g. sensor versions, processing levels, geographic coverage, temporal coverage, quality filters, and known biases). Multimodal pretraining should incorporate physical support (such as a weather grid, lake pixel, SAR backscatter value, and station measurement). 

Next FMs frontier is the boundary between EO FMs and climate FMs. ClimaX, Prithvi WxC, and NeuralGCM show that FMs ideas are also being used for atmospheric variables, forecasting, downscaling, and hybrid physics-ML simulation \citep{nguyen2023climax,schmude2024prithviwxc,kochkov2024neuralgcm}. Environmental monitoring increasingly relies on broad climate and environmental data and requires this bridge because satellite observations, meteorological drivers, hydrology, and field measurements interact in the same decision system. PANGAEA, GEO-Bench-2, and the recent  literature show that evaluation should distinguish modality, resolution, task family, adaptation method, and deployment setting \citep{marsocci2024pangaea,simumba2025geobench2,corley2026stateoftheart}. Future benchmarks should also include regression, uncertainty calibration, sensor transfer and event holdout.

Architecture research should continue toward sensor-aware and metadata-aware models. Wavelength-conditioned encoders, location-time embeddings, multimodal mixture-of-experts, and generative any-to-any systems are important steps. At the same time, efficient transfer matters. Operational users need models that can be adapted with limited labels, documented compute, and reproducible code. Parameter-efficient tuning, distillation, quantization, and open evaluation harnesses will be as important as model scale. Finally, RSFMs should support geospatial reasoning and decisions. Many EO applications allocate scarce resources: field visits, emergency response, restoration funds, inspection effort, or monitoring stations. Future RSFMs should therefore be evaluated by how they improve decisions under uncertainty, not only by how they improve prediction on static benchmarks. This is where FMs, uncertainty quantification, active learning, and reinforcement learning naturally meet. A final challenge is communication. Operational users need to know what a model observed, what it inferred, how uncertain it is, and when it should not be trusted. Vision-language interfaces may help translate model outputs into usable explanations, but the explanation must remain grounded in sensor physics and evaluation evidence.



\clearpage
\printbibliography

\end{document}